\documentclass[letterpaper,twocolumn,10pt]{article}

\usepackage{iccv}
\usepackage{times}

\usepackage{amssymb,amsthm,amsmath,amscd}
\usepackage{epsfig}
\usepackage{subfigure}
\usepackage{algorithmic}
\usepackage{algorithm}
\usepackage{graphicx}
\usepackage{verbatim}

\newtheorem{problem}{Problem}

\theoremstyle{definition}

\DeclareMathOperator{\dist}{dist}

\newcommand{\be}{\begin{equation}}
\newcommand{\ee}{\end{equation}}

\newcommand{\polmat}{\mathbf{A}_c}

\providecommand{\norm}[1]{\left\lVert#1\right\rVert}

\iccvfinalcopy % *** Uncomment this line for the final submission
\def\iccvPaperID{26} % *** Enter the ICCV Paper ID here

\ificcvfinal\fi

\begin{document}
%\footnote{Correspondence author}
\title{Motion Segmentation by SCC on the Hopkins 155 Database~\thanks{This work was supported by NSF grants \#0612608 and \#0915064.}}

\author{Guangliang Chen, Gilad Lerman\\
School of Mathematics, University of Minnesota \\
127 Vincent Hall, 206 Church Street SE, Minneapolis, MN 55455\\
{\small \tt glchen@math.umn.edu, lerman@umn.edu}}

\maketitle

\begin{abstract}
We apply the Spectral Curvature Clustering (SCC) algorithm to a benchmark database of 155 motion sequences, and show that it outperforms all other state-of-the-art methods. The average misclassification rate by SCC is 1.41\% for sequences having two motions and 4.85\% for three motions.
\end{abstract}

\noindent \textbf{Supp.~webpage}: http://www.math.umn.edu/$\sim$lerman/scc/

\section{Introduction}
\emph{Multiframe motion segmentation} is a very important yet
challenging problem in computer vision. Given multiple image frames
of a dynamic scene taken by a (possibly moving) camera, the task is
to segment the point correspondences in those views into different
motions undertaken by the moving objects. A more formal definition
of the problem appears below.
\begin{problem}
Consider a dynamic scene consisting of $K$ rigid-body motions
undertaken by $K$ objects relative to a moving camera. Suppose that
$F$ frames of images have been taken by the camera, and that $N$
feature points $\mathbf{y}_1,\ldots,\mathbf{y}_N \in \mathbb{R}^3$
are detected on the objects. Let $\mathbf{z}_{ij}\in\mathbb{R}^2$ be
the coordinates of the feature point $\mathbf{y}_j$ in the
$i^\text{th}$ image frame for every $1\leq i\leq F$ and $1 \leq j
\leq N$, and form $N$ trajectory vectors: $\mathbf{z}_j =
[\mathbf{z}'_{1j}\: \mathbf{z}'_{2j}\: \ldots\:
\mathbf{z}'_{Fj}]'\in\mathbb{R}^{2F}$. The task is to separate these
trajectories $\mathbf{z}_1,\ldots,\mathbf{z}_N$ into independent
motions undertaken by those objects.
\end{problem}

There has been significant research on this subject over the past few years
(see~\cite{Vidal08multiframe,Tron07benchmark} for a comprehensive
literature review). According to the assumption on the camera model, those
algorithms can be divided into the following two categories:
\begin{enumerate}
\item
\emph{Affine
methods}~\cite{Fischler81RANSAC,Torr98geometricmotion,Sugaya04Geometric,Vidal05gpca,Yan06LSA,Ma07,Vidal08multiframe,spectral_applied}
assume an affine projection model, so that the trajectories
associated with each motion live in an affine subspace of dimension at most three (or a linear subspace of dimension at most four containing the affine subspace). Thus, the motion segmentation problem is equivalent to a subspace clustering problem. State-of-the-art affine
algorithms that have been applied to this problem include Random Sample Consensus
(RANSAC)~\cite{Fischler81RANSAC,Torr98geometricmotion}, Multi-Stage Learning
(MSL)~\cite{Sugaya04Geometric}, Generalized Principal Component Analysis
(GPCA)~\cite{Vidal05gpca,Ma07,Vidal08multiframe}, Local Subspace Affinity
(LSA)~\cite{Yan06LSA}, and Agglomerative Lossy Compression
(ALC)~\cite{Ma07Compression,Rao08ALC}.
\item
\emph{Perspective
methods}~\cite{Hartley04Three-View,Schindler06n-view,Vidal06Two-View,Goh07Unsupervised,Rao08RAS,Chen09KSCC}
assume a perspective projection model under which point
trajectories associated with each moving object lie on a multilinear
variety. However, clustering multilinear varieties is a
challenging task and very limited research has been done in this direction.
\end{enumerate}

%and apply a modified version of the SCC
%algorithm to
%solve the multiframe motion segmentation problem. The regular SCC algorithm is
%initialized by a fixed number $c$ of randomly selected subsets of points that are ideally
%from same clusters. We will first point out a significant
%disadvantage of this random sampling idea by showing that the
%percentage of ``pure'' subsets selected in this way becomes very small as the
%number of clusters or the intrinsic dimension of the clusters grows. Next, we will propose a novel initialization strategy for SCC
%based on nearest neighbors. The foundation of the new initialization
%lies in the assumption that nearest neighbors often lie in the same
%cluster and approximately span the true model.
%We compare their performance using an extensive
%benchmark dataset, the Hopkins 155
%Database~\cite{Tron07benchmark}. %The database consists of 104 checkerboard sequences, 38
%traffic sequences and 13 articulated sequences.
An extensive benchmark for comparing the performance of these algorithms is the Hopkins 155 Database~\cite{Tron07benchmark}. It contains
155 video sequences along with features extracted and tracked in all frames for each sequence, 120 of which have two motions and the rest (35 sequences) consist of three motions.

In this paper we examine the performance of a recent affine method, Spectral
Curvature Clustering (SCC)~\cite{spectral_applied,spectral_theory}, on the Hopkins 155 database and compare it with %other state-of-the-art affine algorithms, in particular %RANSAC~\cite{Fischler81RANSAC,Torr98geometricmotion}, %MSL~\cite{Sugaya04Geometric}, GPCA~\cite{Vidal05gpca,Ma07,Vidal08multiframe}, LSA~\cite{Yan06LSA}, and %ALC~\cite{Ma07Compression}
other affine algorithms that are mentioned above (their results have been reported in \cite{Tron07benchmark,Rao08ALC} and also partly online at http://www.vision.jhu.edu/data/hopkins155/).

Our experiments show that SCC outperforms all the above-mentioned affine algorithms on this benchmark dataset with an average classification error of 1.41\% for two motions and 4.85\% for three motions. In contrast, the smallest average misclassification rate among all other affine methods is 2.40\% for sequences containing two motions and 6.26\% for sequences with three motions, both achieved by ALC~\cite{Rao08ALC}.

%The SCC algorithm has both established theoretical guarantees~\cite{spectral_theory} and careful numerical estimates~\cite{spectral_applied}.

The rest of the paper is organized as follows. We first briefly review the SCC algorithm in Section~\ref{sec:scc_nn}, and then test in Section~\ref{sec:experiments} the SCC algorithm against other common affine methods on the Hopkins 155 database. Finally, Section~\ref{sec:discussions} concludes with a brief discussion.

\section{Review of the SCC algorithm}\label{sec:scc_nn}
%In this section we first quickly review the Spectral Curvature
%Clustering (SCC) algorithm, and then introduce a new strategy for
%its initialization.

%\subsection{Review of The SCC Algorithm}
The SCC algorithm~\cite[Algorithm~2]{spectral_applied} takes as
input a data set $\mathrm{X}=\{\mathbf{x}_1,\ldots,\mathbf{x}_N\}$,
which is sampled from a mixture of affine subspaces in the Euclidean space
$\mathbb{R}^D$ and possibly corrupted with noise and outliers. The
number of the subspaces $K$ and the maximum\footnote{By using only the
maximal dimension we treat all the subspaces to be $d$-dimensional. This
strategy works quite well in many cases, as demonstrated
in~\cite{spectral_applied}.} of their dimensions $d$ should also be
provided by the user. The output of the algorithm is a partition of
the data into $K$ (disjoint) clusters, $\mathrm{X}=\bigcup_{1\leq
k\leq K} \mathrm{C}_k$, representing the affine subspaces.

The initial step of the SCC algorithm is to randomly select from the data $c$ subsets of (distinct) points with a fixed size $d+1$.
Based on these $c$ $(d+1)$-tuples, an affinity matrix $\polmat
\in\mathbb{R}^{N\times c}$ is formed in the following way. Let
$\mathrm{J}_1, \ldots, \mathrm{J}_c$ be the index sets of the $c$ subsets. Then for
each $1\leq r\leq c$ and $1\leq i\leq N$, if $i\in\mathrm{J}_r$, we set $\polmat(i,r)=0$ by default; otherwise, we form the corresponding union $\mathrm{I}:=[i\; \mathrm{J}_r]$ and define
\begin{align} \label{eq:affinity_tensor}
\polmat(i,r) &:= e^{-{c^2_\mathrm{p}(\mathrm{I})/(2\sigma^2)}},
\end{align}
in which $\sigma>0$ is a fixed constant whose automatic choice is explained later, and
$c^2_\mathrm{p}(\mathrm{I})$ is the (squared) polar
curvature~\cite{spectral_applied} of the corresponding $d+2$ points,
$\mathbf{x}_\mathrm{I} :=[\cdots\mathbf{x}_i\cdots]_{i\in \mathrm{I}}$. That
is,
\begin{align} \label{eq:square_polcurv}
c^2_\mathrm{p}(\mathrm{I}) &:=
\max_{j,k\in\mathrm{I}} \norm{\mathbf{x}_j-\mathbf{x}_k}_2^2
\nonumber \\
& \qquad
\cdot \frac{1}{d+2}\;
\sum_{j\in\mathrm{I}}\frac{\det(\mathbf{x}'_\mathrm{I}\cdot
\mathbf{x}_\mathrm{I}+1)}{\prod_{k\in\mathrm{I},k\ne j}\;
 \norm{\mathbf{x}_j-\mathbf{x}_k}_2^2}.
\end{align}

Note that the numerator $\det(\mathbf{x}'_\mathrm{I}\cdot
\mathbf{x}_\mathrm{I}+1)$ is, up to a factor, the (squared) volume of the $(d+1)$-simplex formed by the $d+2$ points $\mathbf{x}_\mathrm{I}$. Therefore, the polar curvature can be thought of as being the volume of the simplex, normalized at each vertex, averaged over the vertices, and then scaled by the diameter of the simplex. When $d+2$ points are sampled from the same subspace, we expect the polar curvature to be close to zero and consequently the affinity close to one. On the other hand, when they are sampled from mixed subspaces, the polar curvature is expected to be large and the affinity close to zero.

The SCC algorithm next forms pairwise weights $\mathbf{W}$ from the
above multi-way affinities:
\begin{equation}\label{eq:pairwise_weights}
\mathbf{W} = \polmat \cdot \polmat',
\end{equation}
and applies spectral clustering~\cite{Ng02} to find $K$ clusters
$\mathrm{C}_1,\ldots,\mathrm{C}_K$.

In order to refine the clusters, SCC then re-samples $c/K$
$(d+1)$-tuples from each of the clusters $\mathrm{C}_k, 1\leq k \leq K$, and re-applies the rest of the
steps. This procedure is repeated until convergence for a best
segmentation, and is referred to as \emph{iterative sampling}
(see~\cite[Sect.~3.1.1]{spectral_applied}).
Its convergence is measured by the
total \emph{orthogonal least squares (OLS)} error of $d$-dimensional affine subspace approximations
$F_1,\ldots, F_K$ to the clusters
$\mathrm{C}_1,\ldots,\mathrm{C}_K$:
\begin{equation}\label{eq:e_OLS}
e^2_\mathrm{OLS}=\sum_{k=1}^{K}\sum_{\mathbf{x}\in
\mathrm{C}_k}\dist^2(\mathbf{x},F_k).
\end{equation}
In situations where the ground truth labels of the data points are
known, we also compute the misclassification rate:
\begin{equation}
e_{\%} = \frac{\text{\# of misclassified points}}{N} \cdot 100\%.
\end{equation}

The parameter $\sigma$ of Eq.~\eqref{eq:affinity_tensor} is automatically selected by SCC at each iteration in the following way.
Let $\mathbf{c}$ denote the vector of all
the $(N-d-1)\cdot c$ squared polar curvatures computed in an arbitrarily fixed iteration. The algorithm applies the following set of candidate values
which represent several scales of the curvatures:
\begin{equation}\label{set:sigmas}
\{\mathbf{c}\left({(N-d-1)\cdot c}/{K^q}\right)\mid q=1,\ldots,d+1\},
\end{equation}
and chooses the one for which the error of Eq.~\eqref{eq:e_OLS} is minimized.
A quantitative derivation of the above selection criterion for $\sigma$ appears
in \cite[Section~3.1.2]{spectral_applied}. It is also demonstrated in \cite{spectral_applied} that SCC will often fail with arbitrary choices of $\sigma$.

%So far we have not discussed how to tune the parameter $\sigma>0$ (see Eq.~\eqref{eq:affinity_tensor}), which sensitively affects clustering. We fix an arbitrary iteration and use $\mathbf{c}$ to denote the vector of all the $(N-d-1)\cdot c$ squared polar curvatures computed in that iteration and sorted in nondecreasing order. We assert that the optimal value of $\sigma$ is within the range of the vector $\mathbf{c}$. We will search in the following set of candidate values which represent several scales of the curvatures:
%\begin{equation}\label{set:sigmas}
%\{\mathbf{c}\left({(N-d-1)\cdot c}/{K^q}\right)\mid q=1,\ldots,d+1\}.
%\end{equation}
%A quantitative derivation of the above candidate set appears in \cite[Section~3.1.2]{spectral_applied}

%We remark that there are other numerical strategies, such as an
%automatic scheme of tuning the parameter $\sigma$, that are used in
%the practical implementation of the SCC algorithm. For their
%details, the reader is referred
%to~\cite[Sect.~3.1]{spectral_applied}.

We present (a simplified version of) the SCC algorithm below (in Algorithm \ref{alg:scc}). We note that the storage requirement of the algorithm is $O(N\cdot(D+c))$, and the total running time is
$O(n_\textrm{s}\cdot (d+1)^2 \cdot D \cdot N\cdot c)$, where $n_\textrm{s}$ is the number of sampling iterations performed (till convergence, typically $O(d)$).
\begin{algorithm}
\caption{Spectral Curvature Clustering (SCC)}
\label{alg:scc}
\begin{algorithmic}[1]
\REQUIRE Data set $\mathrm{X}$, maximal intrinsic
dimension $d$, and number of subspaces $K$ (required);
number of sampled subsets $c$ (default = $100\cdot K$)
\ENSURE $K$ disjoint clusters $\mathrm{C}_1,\ldots,\mathrm{C}_K$.\\
\hspace{-.27in} \textbf{Steps:}\\
\STATE Sample randomly $c$ subsets of $\mathrm{X}$ (with indices
$\mathrm{J}_1,\ldots,\mathrm{J}_c$), each containing $d+1$
distinct points.
\STATE \label{step:columnwise_polcurv_computation} For each sampled
subset $\mathrm{J}_r$, compute the squared polar curvature of it and
each of the remaining $N-d-1$ points in $\mathrm{X}$ by
Eq.~\eqref{eq:square_polcurv}. Sort increasingly these $(N-d-1)\cdot c$
squared curvatures into a vector $\mathbf{c}$.
\STATE \label{step:spectral_clustering}
\textbf{for} $q=1$ to $d+1$ \textbf{do}\\
\begin{itemize}
\item
Form the matrix $\polmat\in\mathbb{R}^{N\times c}$ by setting
$\sigma^2 = \mathbf{c}((N-d-1)\cdot c/K^q)$ in Eq.~\eqref{eq:affinity_tensor},
and estimate the weights $\mathbf{W}$ via
Eq.~\eqref{eq:pairwise_weights}
\item
Apply spectral clustering~\cite{Ng02} to these weights and find a
partition of the data $\mathrm{X}$ into $K$ clusters
\end{itemize}
\textbf{end for}\\
Record the partition $\mathrm{C}_1,\ldots,\mathrm{C}_K$ that has the
smallest total OLS error, i.e., $e^2_{\text{OLS}}$ of
Eq.~\eqref{eq:e_OLS}, for the corresponding $K$ $d$-dimensional affine subspaces.
\STATE Sample $c/K$ subsets of points (of size $d+1$) from each
$\mathrm{C}_k$ found above and repeat
Steps~\ref{step:columnwise_polcurv_computation} and
\ref{step:spectral_clustering} to find $K$ newer clusters. Iterate
until convergence to obtain a best segmentation.
\end{algorithmic}
\end{algorithm}

\begin{figure*}[htbp]
\centering
\includegraphics[width=0.32\textwidth]{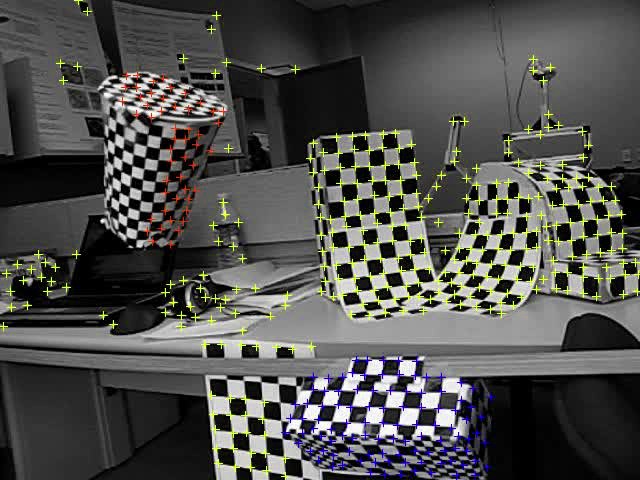}
\includegraphics[width=0.32\textwidth]{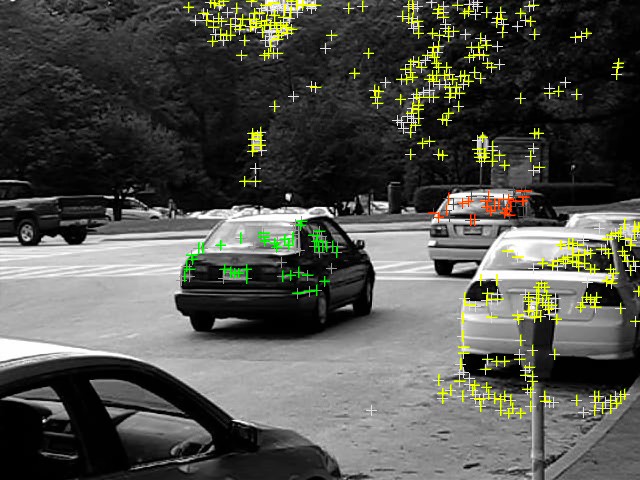}
\includegraphics[width=0.32\textwidth]{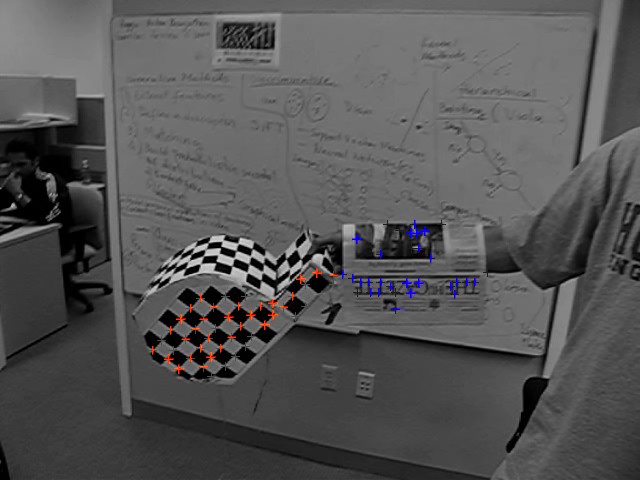}
\caption{A sample image from each of the three categories in the Hopkins155 database.}
\label{fig:sample_images}
\end{figure*}

\section{Results}\label{sec:experiments}
We compare the SCC algorithm with other
state-of-the-art affine methods, such as ALC~\cite{Ma07Compression,Rao08ALC},
GPCA~\cite{Vidal05gpca,Ma07,Vidal08multiframe}, LSA~\cite{Yan06LSA},
MSL~\cite{Sugaya04Geometric}, and RANSAC~\cite{Fischler81RANSAC,Torr98geometricmotion}, using the
Hopkins 155 benchmark~\cite{Tron07benchmark}.
We also compare the performance of affine methods
with an oracle, the Reference algorithm (REF)~\cite{Tron07benchmark}, which fits subspaces using the ground truth clusters and re-assigns points to its nearest subspace. Though it cannot be used in practice, REF verifies the validity of affine camera model and provides a basis for comparison among practical algorithms. The results of the latter six
methods (including REF) are already published in \cite{Vidal08multiframe,Rao08ALC}, so we simply copy them from there.

\begin{table}[htbp]
\centering \caption{\small Summary information of the Hopkins 155 database: number of sequences (\# Seq.), average number of feature points ($N$), and average number of frames ($F$) in each category for two motions and three motions separately.}
\vspace{.1in}
\begin{tabular}{|l||c|c|c||c|c|c|}
  \hline
    &\multicolumn{3}{c||}{2 motions}
    &\multicolumn{3}{c|}{3 motions}\\
    \cline{2-7}
     & \# Seq. & $N$ & $F$ & \# Seq. & $N$ &$F$ \\
  \hline\hline
Checker.   & 78 & 291 & 28 & 26 & 437 & 28\\
Traffic        & 31 & 241 & 30 &  7 & 332 & 31\\
Other          & 11 & 155 & 40 &  2 & 122 & 31\\
All            & 120& 266 & 30 & 35 & 398 & 29\\ \hline
\end{tabular}
\label{tab:hopkins155_description}
\end{table}

The Hopkins 155 database contains sequences with two and three
motions, and consists of three categories of motions (see Figure~\ref{fig:sample_images} for a sample image in each category and Table~\ref{tab:hopkins155_description} for some summary information of each category, e.g., number of sequences, average number of tracked features, and average number of frames):
%\begin{description}
%  \item[Checkerboard:] this category consists of 104 sequences of indoor scenes taken with
%a handheld camera under controlled conditions.
%  \item[Traffic:] this category consists of 38 sequences of outdoor traffic scenes
%taken by a moving handheld camera.
%  \item[Other (Articulated/Non-rigid):]
%  this category contains 13 sequences displaying motions constrained
%by joints, head and face motions, people walking, etc..
%\end{description}
%
\begin{itemize}
\item \textbf{Checkerboard}:
this category consists of 104 sequences of indoor scenes taken with
a handheld camera under controlled conditions. %The checkerboard
%pattern on the objects is used to assure a large number of tracked
%points.
\item \textbf{Traffic}:
this category consists of 38 sequences of outdoor traffic scenes
taken by a moving handheld camera.
\item \textbf{Other (Articulated/Non-rigid)}:
this category contains 13 sequences displaying motions constrained
by joints, head and face motions, people walking, etc.
\end{itemize}
%

%Table~\ref{tab:hopkins155_description} presents some summary information about the
%dataset: the number of sequences, the average number of tracked
%features, and the average number of frames for each one of the
%categories aforementioned, distinguishing also between sequences with
%two and three motions.

It is proved (e.g., in~\cite{Ma07}) that the trajectory vectors
associated with each motion live in a distinct affine subspace of dimension $d\leq 3$ (or a linear subspace of dimension $d\leq 4$ containing the affine subspace). Also, it is possible to cluster the trajectories either in the full space $\mathbb{R}^{2F}$ ($F$ is the number of frames) or in some projected space (after dimensionality reduction by PCA), e.g., $\mathbb{R}^{4K}$ ($K$ is the number of motions) or $\mathbb{R}^{d+1}$. Thus, we will apply the SCC algorithm (Algorithm~\ref{alg:scc}) to each of the 155 motion sequences to segment $d$-dimensional subspaces in $\mathbb{R}^D$ in six ways: $(d,D) = (3,4), (3,4K), (3,2F), (4,5), (4,4K), (4,2F)$. Each case is correspondingly represented by the shorthand \emph{SCC $(d,D)$}.

\begin{table*}[htbp]
\centering \caption{\small Misclassification rates for sequences with \textit{two} motions. ALC 5 and ALC sp respectively represent ALC with projection dimensions 5 and a sparsity-preserving dimension, LSA $n$ means applying LSA in the projected space $\mathbb{R}^n$ (after dimensionality reduction), and REF refers to the reference algorithm.}
\vspace{.1in}
\begin{tabular}{|l||r|r||r|r||r|r||r|r|}
\hline
    &\multicolumn{2}{c||}{Checkerboard}
    &\multicolumn{2}{c||}{Traffic}
    &\multicolumn{2}{c||}{Other}
    &\multicolumn{2}{c|}{All}\\
    \cline{2-9}
    & mean    & median&mean&median&mean&median&mean&median\\
\hline\hline
%SCC 5 \\ SCC 4K \\SCC 2F\\
ALC 5       &2.66\%&0.00\%&2.58\%&0.25\%&6.90\%&0.88\%&3.03\%&0.00\%\\
ALC sp      &1.55\%&0.29\%&1.59\%&1.17\%&10.70\%&0.95\%&2.40\%&0.43\%\\
GPCA        &6.09\%&1.03\%&1.41\%&0.00\%&2.88\%&0.00\%&4.59\%&0.38\%\\
LSA 5       &8.84\%&3.43\%&2.15\%&1.00\%&4.66\%&1.28\%&6.73\%&1.99\%\\
LSA $4K$    &2.57\%&0.27\%&5.43\%&1.48\%&4.10\%&1.22\%&3.45\%&0.59\%\\
MSL         &4.46\%&0.00\%&2.23\%&0.00\%&7.23\%&0.00\%&4.14\%&0.00\%\\
RANSAC      &6.52\%&1.75\%&2.55\%&0.21\%&7.25\%&2.64\%&5.56\%&1.18\%\\
REF         &2.76\%&0.49\%&0.30\%&0.00\%&1.71\%&0.00\%&2.03\%&0.00\%\\
SCC $(3,4)$ &2.99\%&0.39\%&1.20\%&0.32\%&7.71\%&3.67\%&2.96\%&0.42\%\\
SCC $(3,4K)$&1.76\%&0.01\%&0.46\%&0.16\%&4.06\%&1.69\%&1.63\%&0.06\%\\
SCC $(3,2F)$&1.77\%&0.00\%&0.63\%&0.14\%&4.02\%&2.13\%&1.68\%&0.07\%\\
SCC $(4,5)$ &2.31\%&0.25\%&0.71\%&0.26\%&5.05\%&1.08\%&2.15\%&0.27\%\\
SCC $(4,4K)$&1.30\%&0.04\%&1.07\%&0.44\%&3.68\%&0.67\%&1.46\%&0.16\%\\
SCC $(4,2F)$&1.31\%&0.06\%&1.02\%&0.26\%&3.21\%&0.76\%&1.41\%&0.10\%\\
 \hline
\end{tabular}
\label{tab:two_motions}
\end{table*}

\begin{table*}[htbp]
\centering \caption{\small Misclassification rates for sequences with \textit{three} motions. ALC 5 and ALC sp respectively represent ALC with projection dimensions 5 and a sparsity-preserving dimension, LSA $n$ means applying LSA in the projected space $\mathbb{R}^n$ (after dimensionality reduction) and REF refers to the reference algorithm.}
\vspace{.1in}
\begin{tabular}{|l||r|r||r|r||r|r||r|r|}
\hline
    &\multicolumn{2}{c||}{Checkerboard}
    &\multicolumn{2}{c||}{Traffic}
    &\multicolumn{2}{c||}{Other}
    &\multicolumn{2}{c|}{All}\\
    \cline{2-9}
    & mean& med.&mean&med.&mean&med.&mean&med.\\
\hline\hline
%SCC 5 \\ SCC 4K \\SCC 2F\\
ALC 5
&7.05\%&1.02\%&3.52\%&1.15\%&7.25\%&7.25\%&6.26\%&1.02\%\\
ALC sp
&5.20\%&0.67\%&7.75\%&0.49\%&21.08\%&21.08\%&6.69\%&0.67\%\\
GPCA        &31.95\%&32.93\%&19.83\%&19.55\%&16.85\%&16.85\%&28.66\%&28.26\%\\
LSA 5       &30.37\%&31.98\%&27.02\%&34.01\%&23.11\%&23.11\%&29.28\%&31.63\%\\
LSA $4K$    &5.80\%&1.77\%&25.07\%&23.79\%&7.25\%&7.25\%&9.73\%&2.33\%\\
MSL         &10.38\%&4.61\%&1.80\%&0.00\%&2.71\%&2.71\%&8.23\%&1.76\%\\
RANSAC      &25.78\%&26.01\%&12.83\%&11.45\%&21.38\%&21.38\%&22.94\%&22.03\%\\
REF         &6.28\%&5.06\%&1.30\%&0.00\%&2.66\%&2.66\%&5.08\%&2.40\%\\
SCC $(3,4)$ &7.72\%&3.21\%&0.52\%&0.28\%&8.90\%&8.90\%&6.34\%&2.36\%\\
SCC $(3,4K)$&6.00\%&2.22\%&1.78\%&0.42\%&5.65\%&5.65\%&5.14\%&1.67\%\\
SCC $(3,2F)$&6.23\%&1.70\%&1.11\%&1.40\%&5.41\%&5.41\%&5.16\%&1.58\%\\
SCC $(4,5)$ &5.56\%&2.03\%&1.01\%&0.47\%&8.97\%&8.97\%&4.85\%&2.01\%\\
SCC $(4,4K)$&5.68\%&2.96\%&2.35\%&2.07\%&10.94\%&10.94\%&5.31\%&2.40\%\\
SCC $(4,2F)$&6.31\%&1.97\%&3.31\%&3.31\%&9.58\%&9.58\%&5.90\%&1.99\%\\
\hline
\end{tabular}
\label{tab:three_motions}
\end{table*}

We use the default value $c=100\cdot K$ for all SCC $(d,D)$ when applied to the 155 sequences. Also, in order to mitigate the randomness effect due to initial sampling, we repeat the experiment 100 times and record
only the average misclassification rate. For each SCC $(d,D)$, we report in Table~\ref{tab:two_motions} the mean and median of the averaged errors for sequences with two motions, and in Table~\ref{tab:three_motions} results on three motions. Figure~\ref{fig:hists} shows histograms of the misclassification rates with the percentage of sequences in which each algorithm achieved a certain error. The corresponding histograms for other methods are shown in~\cite[Figure 3]{Vidal08multiframe}.

%\begin{figure*}[ht]
%%\centering
%\includegraphics[width=\textwidth]{hist_other.pdf}
%%\includegraphics[width=\textwidth]{hist_3motions.pdf}}
%\caption{Histograms of misclassification errors obtained by other methods (copied from~\cite[Figure 3]{Vidal08multiframe}).}
%\label{fig:hists_other}
%\end{figure*}

\section{Discussion}\label{sec:discussions}

\begin{figure*}[htbp]
\centering
\subfigure[two motions]{
\includegraphics[width=\textwidth]{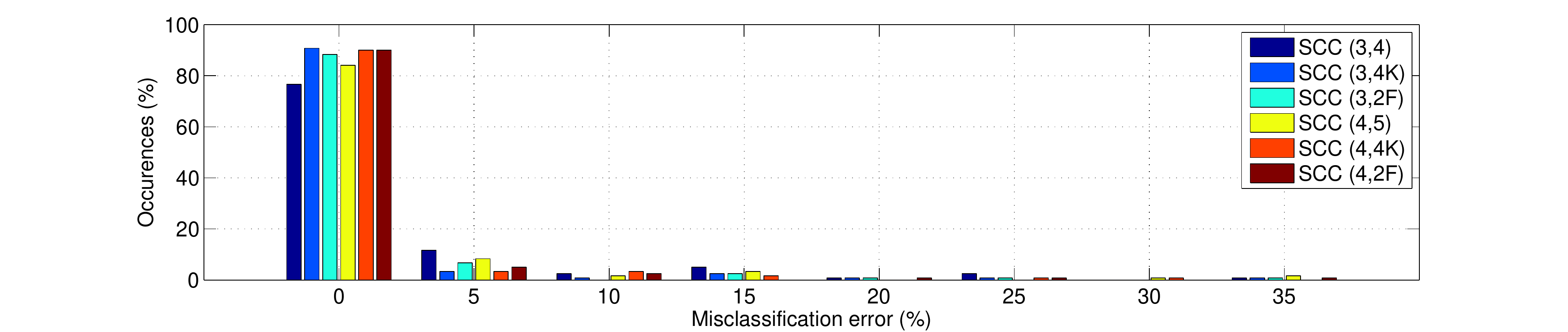}}\\
\subfigure[three motions]{
\includegraphics[width=\textwidth]{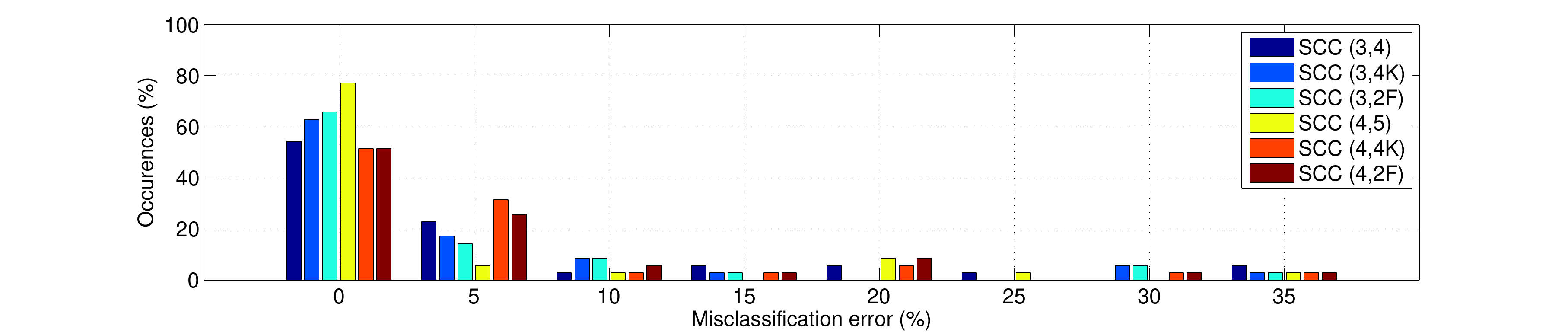}}
\caption{Histograms of misclassification errors obtained by SCC.}
\label{fig:hists}
\end{figure*}

Looking at Tables~\ref{tab:two_motions} and \ref{tab:three_motions}, we conclude that the SCC algorithm (with all six pairs $(d,D)$) outperforms all competing methods (in terms of the mean error) and is very close to the reference algorithm (REF). In the checkerboard category, it even has a better performance than REF. In addition, SCC has the following two strengths in comparison with most other affine methods. First, as we observed in experiments, the performance of SCC is not so sensitive to its free parameter $c$. In contrast, the ALC algorithm is very sensitive to its distortion parameter $\varepsilon$ and often gives incorrect number of clusters, requiring running it for many choices of $\varepsilon$ while having no theoretical guarantee. Second, SCC can be directly applied to the original trajectory vectors (which are very high dimensional), thus preprocessing of the trajectories, i.e., dimensionality reduction, is not necessary (unlike GPCA and LSA). Finally, we remark that SCC also outperforms some perspective methods, e.g., Local Linear Manifold Clustering (LLMC)~\cite{Goh07Unsupervised} (their misclassification rates are also available at http://www.vision.jhu.edu/data/hopkins155/).

The histograms (in Figure~\ref{fig:hists}) show that the SCC algorithm obtains a perfect segmentation for 80\% of two-motion sequences and for over 50\% of three-motion sequences. Under this criterion, SCC is at least comparable to the best algorithms (ALC, LSA $4K$, MSL) and the reference algorithm (REF); see \cite[Figure 4]{Rao08ALC} and \cite[Figure 3]{Vidal08multiframe}. Moveover, SCC has the shortest tails; its worst case segmentation error (about 35\%) is much smaller than those of other methods some of which are as large as 50\%.

Regarding running time, the SCC algorithm generally takes 1 to 2 seconds to process one sequence on a compute server with two dual core AMD Opteron 64-bit 280 processors (2.4 GHz) and 8 GB of RAM. It is much faster than the best competitors such as ALC, LSA $4K$, and MSL (see their computation time in \cite[Table 6]{Rao08ALC} and \cite[Tables 3 \& 5]{Tron07benchmark} while also noting that there were all performed on faster machines).

At the time of finalizing this version we have found out
about the very recent affine method of Sparse Subspace Clustering
(SSC)~\cite{Elhamifar09SSP} which reportedly has superb results on the
Hopkins 155 database and outperforms the results reported
here for both SCC and REF. It will be interesting to test
its sensitivity to its tuning parameter $\lambda$ in future work.

\section*{Acknowledgements}
We thank the anonymous reviewers for their helpful comments and for pointing out reference \cite{Elhamifar09SSP} to us. Special thanks go to Rene Vidal for encouraging comments when this manuscript was still at an early stage and for referring us to the workshop. Thanks to the Institute for Mathematics and its Applications (IMA), in particular Doug Arnold and Fadil Santosa, for an effective hot-topics workshop on multi-manifold modeling that we participated in. The research described in this paper was partially supported by NSF grants \#0612608 and \#0915064.

\bibliographystyle{ieee}

\end{document}